\documentclass[review]{elsarticle}
\usepackage{amsmath}
\usepackage{bm}
\usepackage{diagbox}
\usepackage{booktabs}
\usepackage{lineno,hyperref}
\usepackage{multirow}
\usepackage{bbding}

\journal{}

\begin{document}

\begin{frontmatter}

\title{Video Saliency Prediction Using Enhanced Spatiotemporal Alignment Network}
\author[]{Jin Chen\textsuperscript{\rm 1}}
\author[]{Huihui Song\textsuperscript{\rm 1}}
\author[]{Kaihua Zhang\corref{cor1}\textsuperscript{\rm 1}}
\ead{zhkhua@gmail.com}
\author[]{Bo Liu\textsuperscript{\rm 2}}
\cortext[cor1]{Corresponding author. Phone number: 086-13851581017}
\author[]{Qingshan Liu\textsuperscript{\rm 1}}

\address{\textsuperscript{\rm 1}B-DAT,CICAEET, Nanjing University of Information Science and Technology, Nanjing, China\\
\textsuperscript{\rm 2}JD Finance America Corporation}
\begin{abstract}
%
%Video saliency prediction aims to estimate human's dynamic eye-fixation maps,
%which has been widely applied into numerous computer vision tasks.
%
Due to a variety of motions across different frames, it is highly challenging to learn an effective spatiotemporal representation for accurate video saliency prediction (VSP).
To address this issue, we develop an effective spatiotemporal feature alignment network tailored to VSP, mainly including two key sub-networks: a multi-scale deformable convolutional alignment network (MDAN) and a bidirectional convolutional Long Short-Term Memory (Bi-ConvLSTM) network.
The MDAN learns to align the features of the neighboring frames to the reference one in a coarse-to-fine manner, which can well handle various motions.
Specifically, the MDAN owns a pyramidal feature hierarchy structure that first leverages deformable convolution (Dconv) to align the lower-resolution features across frames, and then aggregates the aligned features to align the higher-resolution features, progressively enhancing the features from top to bottom.
The output of MDAN is then fed into the Bi-ConvLSTM for further enhancement, which captures the useful long-time temporal information along forward and backward timing directions to effectively guide attention orientation shift prediction under complex scene transformation.
Finally, the enhanced features are decoded to generate the predicted saliency map.
The proposed model is trained end-to-end without any intricate post processing.
Extensive evaluations on four VSP benchmark datasets demonstrate that the proposed method achieves favorable performance against state-of-the-art methods.
The source codes and all the results will be released at \url{https://github.com/cj4L/ESAN-VSP}.
\end{abstract}


\end{frontmatter}

\section{Introduction}
The objective of VSP is to faithfully model the human's gaze eye-fixation when watching a dynamic scene.
As a branch of object saliency detection, VSP contributes to the cognitive research of human vision attention through understanding and analyzing dynamic video frames.
VSP has been widely used to assist various computer vision applications, such as autonomous driving~\cite{deng2016does}, object detection and recognition~\cite{aguilar2017pedestrian,flores2019saliency}, video segmentation~\cite{wang2017saliency,hu2018unsupervised}, visual tracking~\cite{zhang2019non}, video captioning~\cite{gao2017video}, human-robot interaction~\cite{schillaci2013evaluating} and video summarization~\cite{marat2007video}, to name a few.

In recent years, benefiting from the breakthrough of deep learning (DL), a variety of DL-based VSP methods ~\cite{huang2015salicon,liu2015predicting,pan2016shallow,kruthiventi2017deepfix,wang2017deep} have been proposed to predict eye-fixation allocation in each frame.
Those DL-based techniques take advantage of a large amount of labeled eye-tracking data to learn an effective semantic feature representation in an end-to-end manner that can accurately predict the salient object locations, greatly outperforming the traditional methods~\cite{itti2003realistic,zhang2009sunday,ren2013regularized,yan2018unsupervised} with hand-crafted features.
Different from the static image saliency detection that only needs to consider spatial cues in one image, VSP should also take into account the temporal cues to handle challenging motion scenarios across frames.
The human eye's fixation mechanism is affected by subjective consciousness, which shifts the target of attention as the scene changes caused by camera motion, light change, scene scaling and fast target movement, \textit{etc}.
Complex motions from background and inconsistent foreground patterns together result in the difficulty of VSP.
Hence, how to learn an effective spatiotemporal feature representation that can well guide attention orientation shift prediction under various scene transitions plays a key role in VSP.

Existing approaches for VSP~\cite{bak2017spatio,jiang2017predicting,lai2019video,lai2019video} often explicitly estimate optical-flow field between the reference frame and its adjacent frames to capture short-term temporal information and then simply fuse the temporal and spatial information to complement their characteristics.
STSConvNet~\cite{bak2017spatio} extracts temporal information using optical flow between consecutive video frames and investigates different ways to integrate spatial and temporal cues within a deep two-stream spatiotemporal network architecture for VSP.
OM-CNN~\cite{jiang2017predicting} leverages a CNN-based optical-flow estimation method to measure the motion intensity in all frames to solve dynamic consistence restriction.
STRA-Net~\cite{lai2019video} makes use of two parallel DNN streams to extract the spatial and temporal cues with optical flows as input.
%
%Although thanks to DeepFlow~\cite{weinzaepfel2013deepflow}, FlowNet~\cite{dosovitskiy2015flownet}, FlowNet 2.0~\cite{ilg2017flownet} and other CNN based optical flow predictive method could accelerate the calculation of optical flow.
%
%It is still a time-consuming, complex pre-processing procedure and may take up a lot of hard disk space storage.
%The optical flows only capture the short-term temporal information between two sequential frames, which cannot well handle the drastic appearance variations caused by fast motion or occlusions between two sequential frames.
%
Besides, the aforementioned methods further leverage LSTM or Gated Recurrent Unit (GRU) to capature the long-term temporal information across frames to learn effective spatiotemporal feature representations for VSP.
OM-CNN~\cite{jiang2017predicting} designs a 2C-LSTM architecture to learn temporal correlation of high-dimensional features for VSP.
ACLNet~\cite{wang2018revisiting} presents an attentive CNN-LSTM mechanism to predict human gaze, and encodes static attention to learn a dynamic salient representation by using frame-wise image saliency maps.
SalEMA~\cite{linardos2019simple} extends an image saliency structure to VSP by integrating a ConvLSTM module and wrapping a convolutional layer with a temporal exponential moving average.
STRA-Net~\cite{lai2019video} develops a spatiotemporal residual attentive network that leverages convolutional GRUs to model the attention transitions across video frames.
Despite demonstrated success of widely applying LSTM or GRU to VSP, all the aforementioned methods only leverage forward sequence modeling that only captures the forward-frame information, while omitting the useful backward-frame cues that are also helpful to enhance the spatiotemporal feature representations.
To address this issue, we design the Bi-ConvLSTM that makes full use of the forward and backward frame cues to learn a robust feature representation.
%tackle long-time temporal characteristics modeling problem from forward and reverse video timing directions.

%
However, directly using LSTM or GRU to learn spatiotemporal representations from the simply fused short-term spatiotemporal cues cannot work well when the attention target appearances across frames suffer from severe distortions due to large and complex motions.
An effective approach to address this issue is to employ multi-frame alignment technique to enhance feature representation, which has been widely applied in video super-resolution~\cite{jo2018deep,tian2018tdan,wang2019edvr} for motion compensation.
However, we have not found any work that applies feature alignment technique to VSP.
%
%
%Another branch of alignment studies obtain implicit motion compensation by dynamic filtering~\cite{jo2018deep} or deformable convolution~\cite{tian2018tdan,wang2019edvr} in other computer vision fields.
%
%By dynamically predicting offsets of sampling convolution kernels, deep semantic features from both the reference frame and each supporting frame can be made full use to learn the scene motion pattern.
%
To this end, we design a novel spatiotemporal alignment network to implement feature alignment between the reference frame and its adjacent frame via Dconv~\cite{zhu2019deformable}.
The aligned features are then fed into the Bi-ConvLSTM to learn a robust spatiotemporal feature representation for VSP.
%Differ from ~\cite{tian2018tdan,wang2019edvr}, we further use a modulation mechanism that expands the scope of deformation modeling by learning an additional mask branch.
%

In summary, our main contributions are summarized into threefold:

\begin{itemize}
\item A Muti-scale Deformable convolutional Alignment Network (MDAN) is designed to align the features across frames with the help of Dconv~\cite{zhu2019deformable}. To the best of our knowledge, this is the first work to apply Dconv to VSP.
\item A novel Bi-ConvLSTM is introduced to effectively model the long-term attention shift across video frames, which makes full use of the long-term temporal context information in the forward and backward time directions.
\item Extensive evaluations on four VSP benchmarks including DHF1K~\cite{wang2018revisiting}, HollyWood2~\cite{mathe2014actions}, UCF-sports~\cite{mathe2014actions} and DIEM~\cite{mital2011clustering} demonstrate the proposed method  achieves competing performance against state-of-the-art methods.
\end{itemize}
\section{Related Work}
\subsection{Computational Models for VSP}
Existing VSP methods could be roughly grouped into two categories including static models~\cite{vig2014large,kruthiventi2017deepfix,pan2016shallow,huang2015salicon,liu2016learning,wang2017deep} and dynamic ones~\cite{pan2017salgan,wang2018revisiting,min2019tased,lai2019video}.
%
%Here, we pay more attention to the DL-based visual attention models in recent years.
%
%Recognitive theories of human visual attention boost the early development of static attention models.
%
With the help of large-scale eye-tracking labeled datasets for training, numerous DL-based static saliency models for VSP~\cite{vig2014large,kruthiventi2017deepfix,pan2016shallow,huang2015salicon,liu2016learning,wang2017deep} have been proposed and achieved remarkable performance boosting compared to the traditional approaches.
eDN~\cite{vig2014large} follows an entirely automatic data-driven approach to perform a large-scale search for an optimal ensemble of deep CNN features, and then trains an SVM classifier to predict the saliency maps.
DeepFix~\cite{kruthiventi2017deepfix}, DeepNet~\cite{pan2016shallow} and SALICON~\cite{huang2015salicon} leverage large-scale eye-tracking data to fine-tune the classical image classification networks to generate the corresponding eye-fixation maps.
Mr-CNN~\cite{liu2016learning} employs a multi-resolution CNN guided by both bottom-up visual saliency and top-down visual cues to predict visual saliency.
DVA~\cite{wang2017deep} is based on a skip-layer network structure, which estimates eye-fixation from multiple convolutional layers with various reception fields.

Another research branch of VSP focuses on simulating eye fixation behavior in dynamic scenes~\cite{pan2017salgan,wang2018revisiting,min2019tased,lai2019video}.
The traditional dynamic approaches~\cite{itti2003realistic,zhang2009sunday,ren2013regularized} leverage hand-crafted spatiotemporal features to model visual saliency, which cannot capture rich semantic information from the attention targets that is essential for accurate VSP.
To address this issue, numerous DL-based dynamic saliency models~\cite{pan2017salgan,wang2018revisiting,min2019tased,lai2019video} have been developed with promising performance.
SalGan~\cite{pan2017salgan} proposes a data-driven metric based VSP method that is trained with an adversarial loss function, yielding saliency maps that resemble the ground-truth.
ACLNet~\cite{wang2018revisiting} releases a benchmark dataset for predicting human eye movements during dynamic scene free viewing and proposes a CNN-LSTM network with an attention mechanism for VSP.
SalEMA~\cite{linardos2019simple} introduces a conceptually simple exponential moving average of an internal convolutional state to modify existing network architectures for VSP.
TASED-Net~\cite{min2019tased} designs a 3D fully-convolutional network structure and decodes the encoded features spatially while aggregating all the temporal information for VSP.
STRA-Net~\cite{lai2019video} develops a residual attentive learning network architecture, which enhances the spatiotemporal features by a composite attention mechanism for VSP.
\subsection{Deformable Convolutional Networks}
The deformable convolutional network (DCN) proposed by~\cite{dai2017deformable} aims to enhance the capability of regular convolutions by learning additional offsets from its local neighborhood, and allows the network to adaptively capture more contextual information in a larger receptive field.
DCNv2~\cite{zhu2019deformable} reformulates Dconv and introduces a modulation mechanism that expands the scope of deformation modeling through a more comprehensive integration of deformable convolution within the network.
The superior performance of Dconv has been demonstrated in some other computer vision tasks including video super-resolution~\cite{tian2018tdan,wang2019edvr}, object detection~\cite{yang2019reppoints}, image classification~\cite{zhu2018deformable} and crowd understanding~\cite{liu2019adcrowdnet}.
TDAN and EDVR~\cite{tian2018tdan,wang2019edvr} use Dconv to align features between the reference frame and its corresponding supporting frames for motion compensation in video restoration task.
RepPoints~\cite{yang2019reppoints} leverages Dconv to develop a flexible object representation for accurate geometric localization as well as semantic feature extraction.
DHCNet~\cite{zhu2018deformable} achieves better classification performance for hyperspectral image classification by applying the regular convolutions on the Dconv feature maps.
ADCrowdNet~\cite{liu2019adcrowdnet} designs an attention-injective Dconv to address the accuracy degradation issue in highly-congested noisy scenes for crowd understanding task.

%network figure
\begin{figure*}[t]
\centering
\includegraphics[width=0.9\textwidth]{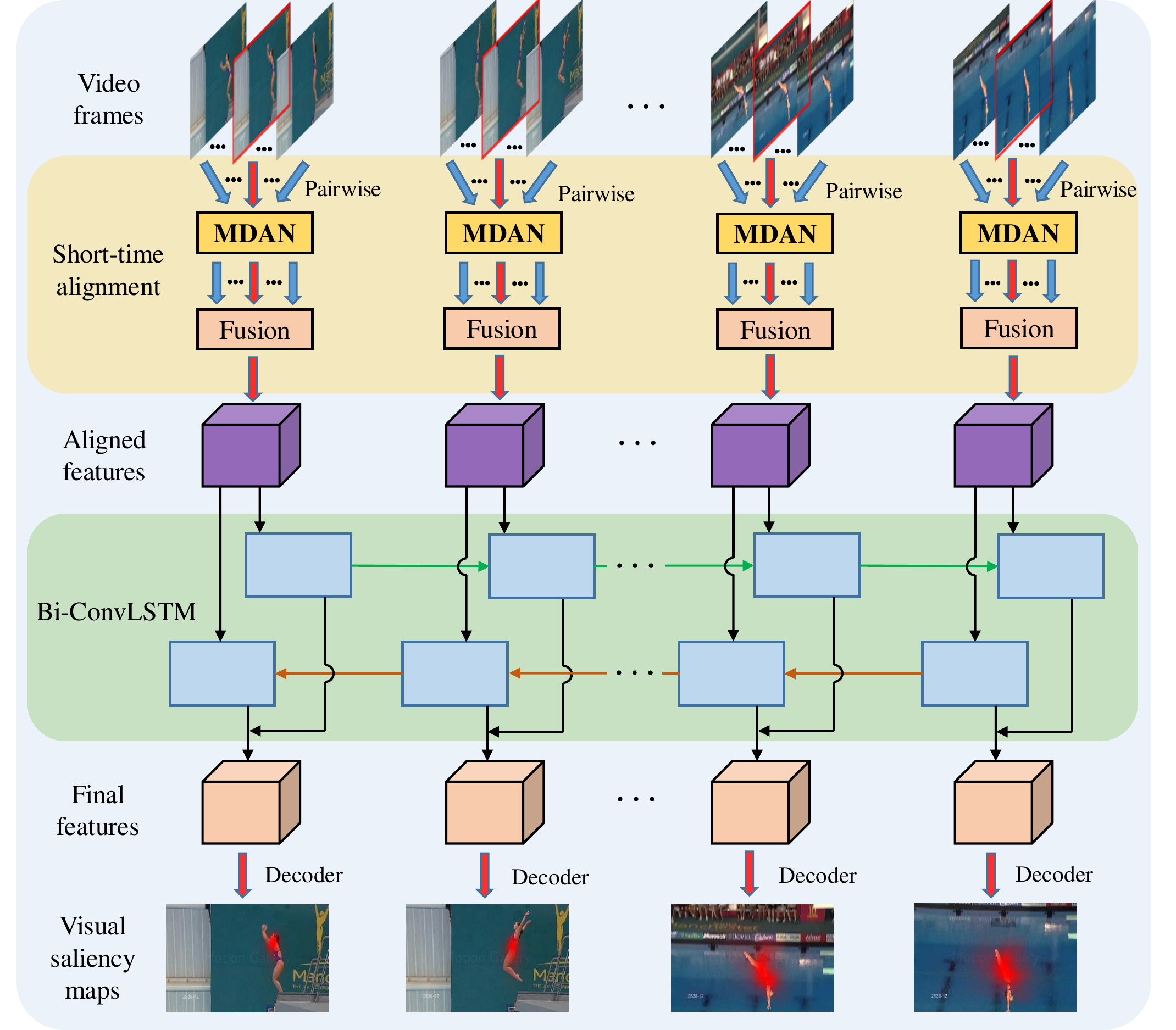}
\caption{Overview of the proposed VSP framework. Firstly, a sequence of video frames are fed into the MDAN to align the reference and its neighboring frames, yielding the aligned features that can well capture short-term spatiotemporal information. Then, the aligned features are sent to the Bi-ConvLSTM, generating the features that encodes long-term spatiotemporal information. Finally, the output features are decoded to generate the predicted saliency map.}
\label{network}
\end{figure*}
%DAM figure
\section{Proposed Approach}

\subsection{Architecture Overview}\label{sec:overview}
Figure~\ref{network} shows an overview of the proposed model for VSP.
Given $N$ consecutive frames $\mathcal{I}=\{\bm{I}_t\}_{t=1}^N$ in a video sequence, our aim is to learn a deep CNN $f_{\mathrm{CNN}}$ that outputs a set of corresponding visual saliency maps $\mathcal{M}=\{\bm{M}_t\}_{t=1}^N$:
\begin{equation}
\mathcal{M}=f_{\mathrm{CNN}}(\mathcal{I};\bm{\theta}_{\mathrm{CNN}}),
\end{equation}
where $\bm{\theta}_{\mathrm{CNN}}$ denotes the whole network parameters to be optimized.
$f_{\mathrm{CNN}}$ consists of three sub-networks: the MDAN $f_{\mathrm{align}}$, the Bi-ConvLSTM $f_{\mathrm{Bi-ConvLSTM}}$ and the decoder $f_{\mathrm{decoder}}$.
Specifically, the design of $f_{\mathrm{align}}$ is inspired by the renowned DCN~\cite{dai2017deformable,zhu2019deformable}. We align each neighboring frame to the reference one at feature level by progressively aligning and aggregating the multi-level features from top to down.
By fusing the spatiotemporal cues across frames at different semantic levels, $f_{\mathrm{align}}$ enables to well handle diverse motions across frames that can severely affect accurately predicting attention shifts in VSP.
Given a sequential of $2T+1$ frames with the reference frame $t$ at the center, $f_{\mathrm{align}}$ aligns the left-and the right-side $T$ neighboring frames to the reference frame $t$, respectively and then fuses them to generate the enhanced reference frame features ${\bm{A}}_t$:
\begin{equation}
\label{eq:align}
{\bm{A}}_t=f_{\mathrm{align}}(\{I_n\}_{n={t-T}}^{t+T};\bm{\theta}_{\mathrm{align}}),
\end{equation}
where $\bm{\theta}_{\mathrm{align}}$ denotes the corresponding network parameters of MDAN to be optimized.
%
%By fusing spatiotemporal information across frames at different semantic levels, $f_{\mathrm{align}}$ can well handle diverse motions across frames that can severely affect accurately predicting attention shifts across frames.

Although the features ${\bm{A}}_t$ in (\ref{eq:align}) are strengthen by the features of the neighbouring frames, their representative capability will be severely affected when the attention targets suffer from severe distortions caused by long-term occlusions or large motions.
To address this issue, we further design the Bi-ConvLSTM $f_{\mathrm{Bi-ConvLSTM}}$ to maintain long-term visual attention stability, generating the enhanced spatiotemporal representation as:
\begin{equation}
\label{eq:biconvlstm}
\{\bm{H}_t^f,\bm{H}_t^b\}=f_{\mathrm{Bi-ConvLSTM}}(\{{\bm{A}}_t, \bm{H}_{t-1}^f, \bm{H}_{t+1}^b\};\bm{\theta}_{\mathrm{Bi-ConvLSTM}}),
\end{equation}
where $\bm{H}_t^f$ and $\bm{H}_t^b$ denotes the forwardly and backwardly estimated hidden states for frame $t$, respectively.
\begin{figure*}[t]
\centering
\includegraphics[width=1\textwidth]{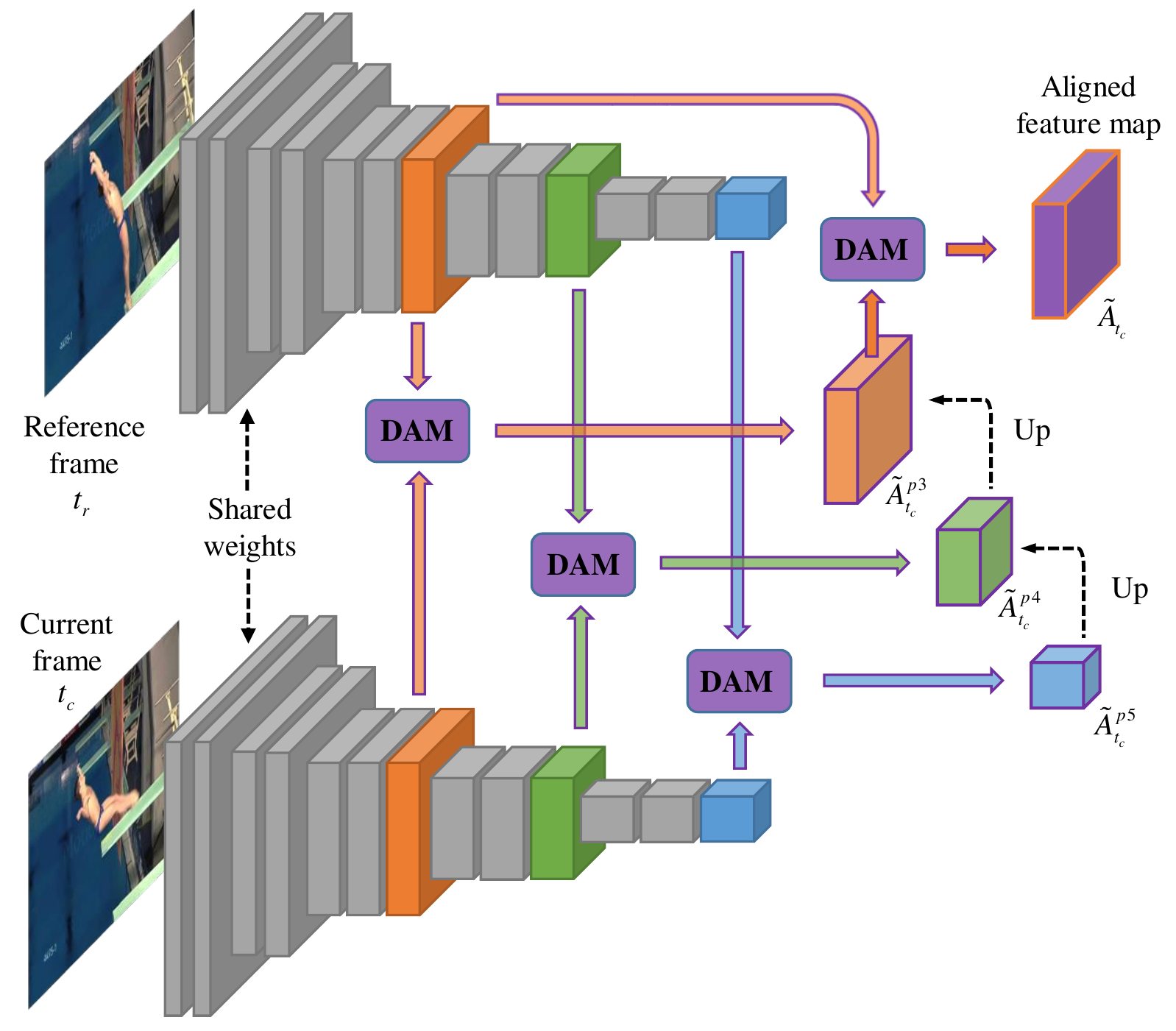}
\caption{Architecture of MDAN.}
\label{MDAN}
\end{figure*}

Finally, the spatiotemporal representations $\{\bm{H}_t^f,\bm{H}_t^b\}$ in (\ref{eq:biconvlstm}) are fused to $\bm{Y}_t$ using (\ref{eq:bi_convlstm}) and fed into the decoder network $f_{\mathrm{decoder}}$ that is composed of a few convolutional layers and a bilinear upsamping layer, yielding the finally predicted saliency map of frame $t$:
\begin{equation}
\label{eq:encoder}
\textit{\textbf{M}}_t = f_{\mathrm{decoder}}(\bm{Y}_t;\bm{\theta}_{\mathrm{decoder}}),
\end{equation}
where $\bm\theta_{\mathrm{decoder}}$ is the parameters of the decoder sub-network to be learned.
\begin{figure*}[t]
\centering
\includegraphics[width=1\textwidth]{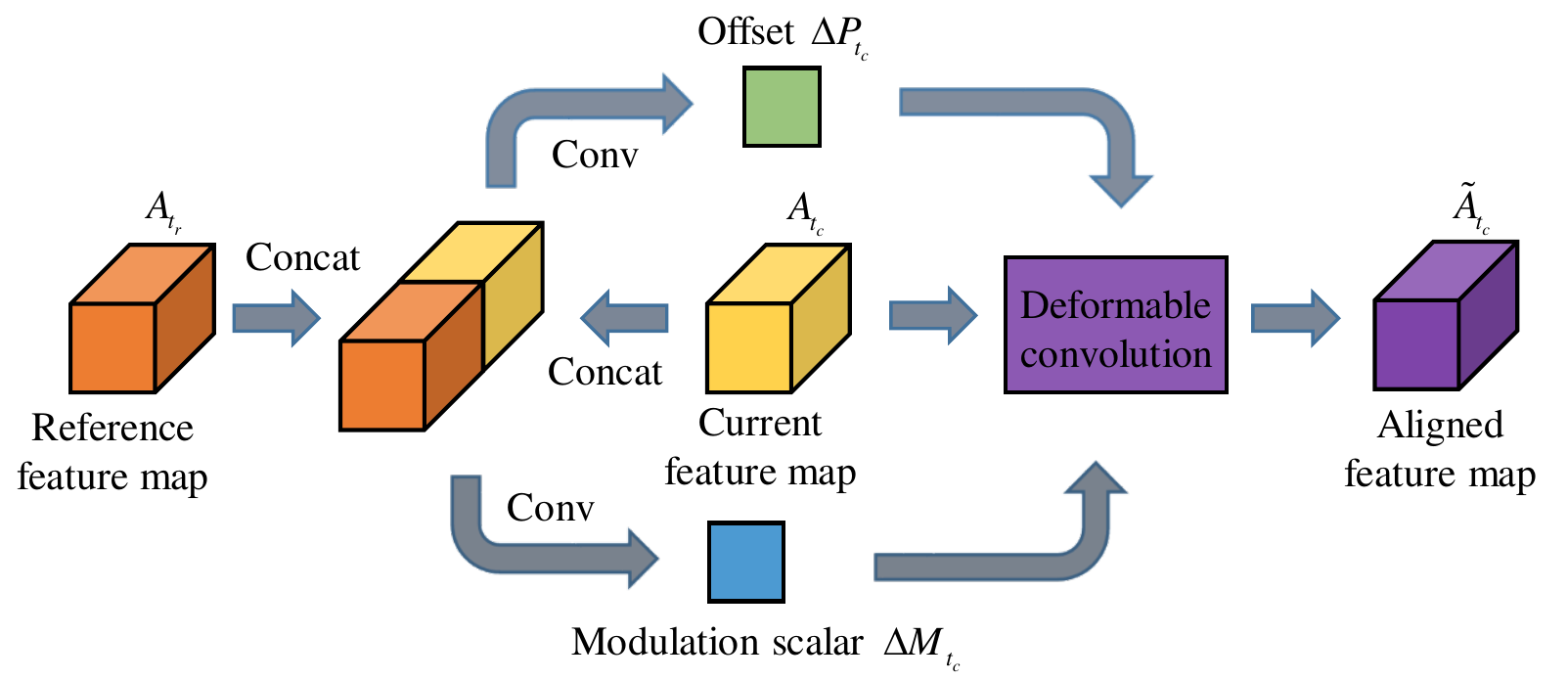}
\caption{Architecture of DAM.}
\label{DAM}
\end{figure*}
%MDAN figure
%\subsection{Deformable Convolutional Alignment Network (DAN)}
\subsection{Multi-scale Deformable Convolutional Alignment Network (MDAN)}
Figure~\ref{MDAN} illustrates the architecture of MDAN, which progressively makes feature alignment in a coarse-to-fine manner through a set of deformable convolutional alignment modules (DAMs).
MDAN can well capture large and complex motion information by adaptively sampling at multiple feature levels in a coarse-to-fine fashion, and does not need to explicitly estimate the motion fields as optical flow~\cite{bak2017spatio,jiang2017predicting,lai2019video,lai2019video}, thereby greatly reducing computational cost.
%
%Next, we first introduce the DAM in details and then

%
Figure~\ref{DAM} shows the architecture of DAM that is based on Dconv~\cite{zhu2019deformable}.
%
%Next, to comprehensively introduce the technique details of DAM, we first briefly retrospect the principle of deformable convolution~\cite{zhu2019deformable}.
%
In~\cite{zhu2019deformable}, the Dconv that maps the input feature $\textit{\textbf{A}}$ to the output features maps $\widetilde{\textit{\textbf{A}}}$ is defined as
\begin{equation}
\label{eq:dcnv2}
\widetilde{\textit{\textbf{A}}}(\textit{\textbf{p}}) = \sum\limits_{k=1}^{K} \textit{\textbf{w}}_{k} \cdot \textit{\textbf{A}}(\textit{\textbf{p}}+\textit{\textbf{p}}_{k}+\Delta \textit{\textbf{p}}_{k}) \cdot \Delta \textit{\textbf{m}}_{k},
\end{equation}
where $K$ denotes the number of sampling locations in a convolutional kernel. For instance, if $K=9$, $\textit{\textbf{p}}_{k}\in \{(-1,-1),(-1,0),\ldots,(1,1)\}$ defines a $3\times 3$ convolutional kernel of dilation 1. $\Delta \textit{\textbf{p}}_{k}$ and $\Delta \textit{\textbf{m}}_{k}$ denote the learnable offset and modulation scalar at the $k$-th location, respectively. $\textit{\textbf{w}}_k$ and $\textit{\textbf{p}}_k$ denote the weight and pre-specified offset of the $k$-th position, respectively

%where $x(p)$ and $y(p)$ denote the features at location $p$ from the input feature maps $x$ and output feature maps $y$, and $w_{k}$ and $p_{k}$ is the weight and pre-specified offset for the $k$-th location respectively.
%%
%With the idea of modulation networks, $\Delta p_{k}$ is a real number with unconstrained range, and the value range of $\Delta m_{k}$ is $\left[0, 1 \right]$.

We employ the Dconv (\ref{eq:dcnv2}) for feature alignment. Given the feature maps $\textit{\textbf{A}}_{t_r}$ and $\textit{\textbf{A}}_{t_c}$ at the reference frame $t_r$ and the current frame $t_c$, respectively, we concatenate the features $[\textit{\textbf{A}}_{t_r},\textit{\textbf{A}}_{t_c}]$ as input, and learn the offset and modulation scalar by
\begin{equation}
\begin{cases}
\Delta\textit{\textbf{P}}_{t_c}=f_{\mathrm{offset}}([\textit{\textbf{A}}_{t_r},\textit{\textbf{A}}_{t_c}];\bm{\theta}_{\mathrm{offset}}),\\
\Delta\textit{\textbf{M}}_{t_c}=f_{\mathrm{modulation}}([\textit{\textbf{A}}_{t_r},\textit{\textbf{A}}_{t_c}];\bm{\theta}_{\mathrm{modulation}}),
\end{cases}
\label{eq:modulation}
\end{equation}
where $\Delta\textit{\textbf{P}}=\{\Delta\textit{\textbf{p}}\}$ and $\Delta\textit{\textbf{M}}=\{\Delta\textit{\textbf{m}}\}$. $f_{\mathrm{offset}}$ and $f_{\mathrm{modulation}}$ are two networks consisting a few convolution layers with parameters $\bm{\theta}_{\mathrm{offset}}$ and $\bm{\theta}_{\mathrm{modulation}}$, respectively.
Afterwards, we replace $\textit{\textbf{A}}$ in (\ref{eq:dcnv2}) by $\textit{\textbf{A}}_{t_c}$, generating the corresponding aligned feature maps $\widetilde{\textit{\textbf{A}}}_{t_c}$:
\begin{equation}
\label{eq:DAM}
\widetilde{\textit{\textbf{A}}}_{t_c}(\textit{\textbf{p}}) = \sum\limits_{k=1}^{K} \textit{\textbf{w}}_{k} \cdot \textit{\textbf{A}}_{t_c}(\textit{\textbf{p}}+\textit{\textbf{p}}_{k}+\Delta \textit{\textbf{p}}_{t_ck}) \cdot \Delta \textit{\textbf{m}}_{t_ck},
\end{equation}
where $\Delta\textit{\textbf{p}}_{t_c}\in \Delta\textit{\textbf{P}}_{t_c}$ and $\Delta\textit{\textbf{m}}_{t_c}\in \Delta\textit{\textbf{M}}_{t_c}$ in (\ref{eq:modulation}).
%
%Then, we introduce the use of deformable convolution for alignment~\cite{tian2018tdan}: align features of each adjacent frame to the reference frame.
%%
%Not like optical-flow based methods to calculate the offset field between all pixels, deformable alignment is employed on features of each frame, denoted by: $F_{t+i}, i \in \left[-T, +T \right]$.
%%
%Concatenate the features of reference frame and the a neighbouring one:
%%
%\begin{equation}
%\label{eq:deformable}
%\Delta P_{t+i} = f(\left[F_{t+i},F{t}\right]),  i \in \left[-T, +T \right]
%\end{equation}
%%
%where  $\Delta P = \left\{ \Delta p \right\}$, $\left[\cdot , \cdot \right]$ is the concatenation operation, and $f$ denote a general function containing few convolution layers.
%%
%The detailed structure of DAM is shown in Figure \ref{DAM}, by concatenating current frame feature map and reference frame feature map, the fusion feature are then sent two convolution branches for learning offset and mask respectively, finally applying the learned offset and mask to the current frame feature map to gain the aligned feature map by deformable convolution.
%\subsection{Multi-scale Deformable Convolutional Alignment Network (MDAN)}
%

%Figure~\ref{MDAN} illustrates the architecture of MDAN.
%
%We design MDAN to address the complex motions, scene and parallax transformation, instead of the theory in optical flow.
%

%Thus, the computational cost is greatly reduced without affecting the alignment accuracy.
%
After introducing the DAM, we give the details of how to design the MDAN.
As shown in Figure \ref{MDAN}, given the reference frame $t_r$ and the current frame $t_c$ as input, we utilize a Siamese network architecture with the VGG16~\cite{simonyan2014very} backbone network to extract their features.
%
%Specifically, we first select three different-level features $p3$, $p4$ and $p5$ corresponding to $pool3$, $pool4$ and $pool5$ layers in VGG16, respectively.
Specifically, we first select three different-level feature maps $\{\bm{A}_{t_r}^{p3},\bm{A}_{t_r}^{p4},\bm{A}_{t_r}^{p5}\}$ and $\{\bm{A}_{t_c}^{p3},\bm{A}_{t_c}^{p4},\bm{A}_{t_c}^{p5}\}$ of frames $t_r$ and $t_c$, respectively, which correspond to their pool3, pool4 and pool5 layers in VGG16, respectively.
Then, we progressively align the features of frame $t_c$ to those of frame $t_r$ in a coarse-to-fine manner.
First, we feed the pairs $\{\bm{A}_{t_r}^{pi},\bm{A}_{t_c}^{pi}\}$, $i=3,4,5$ into the formula of DAM (\ref{eq:DAM}), outputting the corresponding multi-level aligned features $\widetilde{\bm{A}}_{t_c}^{pi}$, $i=3,4,5$.
Then, we fuse the multi-level aligned features progressively from top to bottom, yielding the enhanced aligned features as
\begin{equation}
\bm{A}_{t_c}=[[\widetilde{\bm{A}}_{t_c}^{p5} \uparrow, \widetilde{\bm{A}}_{t_c}^{p4}]\uparrow,\widetilde{\bm{A}}_{t_c}^{p3}].
\label{eq:multMNADFtr}
\end{equation}
%
%Meanwhile, $\{\bm{A}_{t_r}^{p4}$ and $\bm{A}_{t_c}^{p4}\}$ are also fed into the DAM, yielding the aligned features $\widetilde{\bm{A}}_{t_c}^{p4}$.
%
Afterwards, we put $\bm{A}_{t_c}$ (\ref{eq:multMNADFtr}) and $\bm{A}_{t_r}\leftarrow\bm{A}_{t_r}^{p3}$ into (\ref{eq:DAM}), yielding the output of MDAN $\widetilde{\bm{A}}_{t_c}$ for frame $t_c$.
Finally, we concatenate all the aligned features $\{\widetilde{\bm{A}}_{t_c}\}_{t_c=t-T}^{t+T}$ and fuse them through a $1\times 1$ convolution layer to yield the enhanced spatiotemporal features for frame $t$
\begin{equation}
\label{eq:fusion}
\bm{A}_t = f_{\mathrm{fusion}}([\widetilde{\bm{A}}_{t_c=t-T},\ldots,\widetilde{\bm{A}}_{t_c=t+T}];\bm{\theta}_{\mathrm{fusion}}),
\end{equation}
where $\bm{\theta}_{\mathrm{fusion}}$ denotes the weight parameters of the $1\times 1$ convolutional layer.
%Send the features maps with three resolution scale of reference frame and current frame into DAM, which is introduced in\ref{DAM}, the coarse-to-fine aligned feature in different scale $A^{p_{3}}$, $A^{p_{4}}$, $A^{p_{5}}$,can be obtained by:
%%
%\begin{equation}
%\label{eq:mdan}
%A^{p_{i}} = DAM(p_{i}^{cur}, p_{i}^{ref}),  i \in \left[3, 5 \right]
%\end{equation}
%%
%where $p_{i}^{cur}$ and $p_{i}^{ref}$ are specific level feature maps both reference frame and current frame.
%%
%Then, through the upsampling of different scale feature maps, we could gain the finer feature representation of two adjacent pictures in space.
%%
%Bilinear interpolation is applied to implement the unsampling operation.
%%
%Finally, a subsequent deformable alignment is cascaded to further polish the coarsely alignment features, which can be denoted as:
%%
%\begin{equation}
%\label{eq:mdan2}
%A = DAM(p_{3}^{ref}, fusion(fusion(A^{p_{5}} \uparrow, A^{p_{4}}) \uparrow, A^{p_{3}})),
%\end{equation}
%%
%where $A$ is the final aligned features of current frame, $fusion$ denotes the operation containing concatenation and convolution successively, and $\uparrow$ refers $\times 2$ upsampling.
%
\subsection{Bidirectional ConvLSTM (Bi-ConvLSTM)}
The aforementioned MDAN aligns the features between the reference frame and its left-and right-side neighbouring frames that can be viewed as a short-time bidirectional spatial alignment process.
%
%The features of the current frame are enhanced by the complement rich information of its adjacent frames.
%
However, video sequence may have large scene transformation and attention shift due to long-term occlusions or large motions, resulting in difficulty by only using short-term information from adjacent frames.
We further strengthen the fused features $\bm{A}_t$ (\ref{eq:fusion}) generated by MDAN through encoding long-term information across more frames, and leverage Bi-ConvLSTM to fully capture long-term spatiotemporal context information in bi-directions.
%
%ConvLSTM~\cite{xingjian2015convolutional} is a well-known technology for processing sequential tasks, which can be traced back to natural language processing.
%
%Four gates(forget, input, state and output) and a cell state together compose an LSTM cell.
%
In~\cite{xingjian2015convolutional}, the ConvLSTM is formulated as:
\begin{equation}
\label{al:convlstm}
\begin{cases}
\bm{I}_{t} &= \sigma(\bm{W}_{I}^{A} \ast \bm{A}_{t} + \bm{W}_{I}^{H} \ast \bm{H}_{t-1}),  \\
\bm{F}_{t} &= \sigma(\bm{W}_{F}^{A} \ast \bm{A}_{t} + \bm{W}_{F}^{H} \ast \bm{H}_{t-1}),  \\
\bm{O}_{t} &= \sigma(\bm{W}_{O}^{A} \ast \bm{A}_{t} + \bm{W}_{O}^{H} \ast \bm{H}_{t-1}), \\
\bm{C}_{t} &= \bm{F}_{t} \circ \bm{C}_{t-1} + \bm{I}_{t} \circ tanh(\bm{W}_{C}^{A} \ast \bm{A}_{t} + \bm{W}_{C}^{H} \ast \bm{H}_{t-1}),  \\
\bm{H}_{t} &= \bm{O}_{t} \circ tanh(\bm{C}_{t}),
\end{cases}
\end{equation}
where $\ast$ is the convolution operator, $\circ$ is the Hadamard product, $\sigma$ denotes the Sigmoid function, and $tanh$ denotes the hyperbolic tangent function.
For different learnable parameters $\bm{W}$, we do convolutions with input feature maps $\bm{A}_t$ and hidden state $\bm{H}_{t-1}$, respectively, and then sum them and feed into the Sigmoid function to obtain an input gate $\bm{I}_{t}$, an output gate $\bm{O}_{t}$ and a forget gate $\bm{F}_{t}$.
The memory cell $\bm{C}_{t}$ plays the role of an accumulator of the state information by updating the ratio of memory and forgetting between the current and the previous moments, respectively.
Finally, the hidden state $\bm{H}_{t}$ is generated by pixel-wise multiplying the output gate $\bm{O}_{t}$  by the memory cell $\bm{C}_{t}$ rescaled by a tanh activation function.

The ConvLSTM can capture long-term information from the past frames well, but does not consider the rich information from the future frames that is helpful to further boost the performance of VSP.
To this end, we design Bi-ConvLSTM that captures both forward and backward long-range context information, yielding the final spatiotemporal feature representation for VSP:
\begin{equation}
\label{eq:bi_convlstm}
\bm{Y}_{t} = \bm{H}_{t}^{f} + \bm{H}_{t}^{b},
\end{equation}
where $\bm{H}_{t}^{f}$ and $\bm{H}_{t}^{b}$ denote the hidden states from forward and backward ConvLSTM units.
Afterwards, the output features $\bm{Y}_{t}$ are sent to the decoder network $f_{\mathrm{decoder}}$ (\ref{eq:encoder}) to generate the predicted saliency map $\bm{M}_t$.

\subsection{Loss Function}
We leverage the loss function similar to that proposed by~\cite{huang2015salicon,lai2019video}, which combines four loss terms related to saliency evaluation metrics. The loss function is formulated as:
\begin{equation}
\label{eq:all_loss}
\mathcal{L}(\bm{P},\bm{Q},\bm{G}) = \mathcal{L}_{\mathrm{NSS}}(\bm{P},\bm{Q}) + \mathcal{L}_{\mathrm{SIM}}(\bm{P},\bm{G}) + \mathcal{L}_{\mathrm{CC}}(\bm{P},\bm{G}) + \mathcal{L}_{\mathrm{KL}}(\bm{P},\bm{G}),
\end{equation}
where  $\bm{P}$ denotes the predicted attention map, $\bm{Q}$ is the ground-truth binary fixation map, and $\bm{G}$ indicates the continuous ground-truth attention map.

$\mathcal{L}_{\mathrm{NSS}}$ originates from normalized scanpath saliency (NSS), which is introduced to the visual saliency field as a simple correspondence measure between saliency maps and ground-truth~\cite{peters2005components}.
$\mathcal{L}_{\mathrm{NSS}}$ computes the average normalized saliency at fixated locations:
\begin{equation}
\label{eq:nss}
\mathcal{L}_{\mathrm{NSS}}(\bm{P},\bm{Q}) = -\frac{1}{N} \sum\limits_{i} \frac{\bm{P} - \mu (\bm{P})}{\sigma (\bm{P})} \cdot \bm{Q}_{i},
\end{equation}
where $\mu (\cdot)$ and $\sigma (\cdot)$ denote the mean and the standard deviation, respectively, $N$ is the number of positive pixels belonging to $\bm{Q}$.

$\mathcal{L}_{\mathrm{SIM}}$ is derived from the similarity (SIM) metric, which measures the similarity between two distributions and computes the sum of the minimum values at each pixel:
\begin{equation}
\label{eq:sim}
\mathcal{L}_{\mathrm{SIM}}(\bm{P},\bm{G}) = - \sum\limits_{i} \min(\bm{P}_{i}, \bm{G}_{i}),
\end{equation}
where $\bm{P}$ and $\bm{Q}$ are normalized to $\sum\limits_{i} \bm{P}_{i} = \sum\limits_{i} \bm{G}_{i} = 1$.

$\mathcal{L}_{\mathrm{CC}}$  measures the correlation or dependence of two variables by linear correlation coefficient (CC):
\begin{equation}
\label{eq:cc}
\mathcal{L}_{\mathrm{CC}}(\bm{P},\bm{G}) = - \frac{cov(\bm{P},\bm{G})}{\sigma (\bm{P}) \times \sigma (\bm{G})},
\end{equation}
where $cov()$ means the covariance and $\sigma (\cdot)$ is the standard deviation.

$\mathcal{L}_{\mathrm{KL}}$ is from Kullback-Leibler (KL) divergence metric, which measures the difference between two probability distributions:
\begin{equation}
\label{eq:kl}
\mathcal{L}_{\mathrm{KL}}(\bm{P},\bm{G}) = \sum\limits_{i} \bm{G}_{i} \log(\frac{\bm{G}_{i}}{\bm{P}_{i}}).
\end{equation}
%
%While NSS, SIM and CC are similarity metrics, we use their negatives as loss terms.

%results figure
\begin{figure*}[t]
\centering
\includegraphics[width=0.95\textwidth]{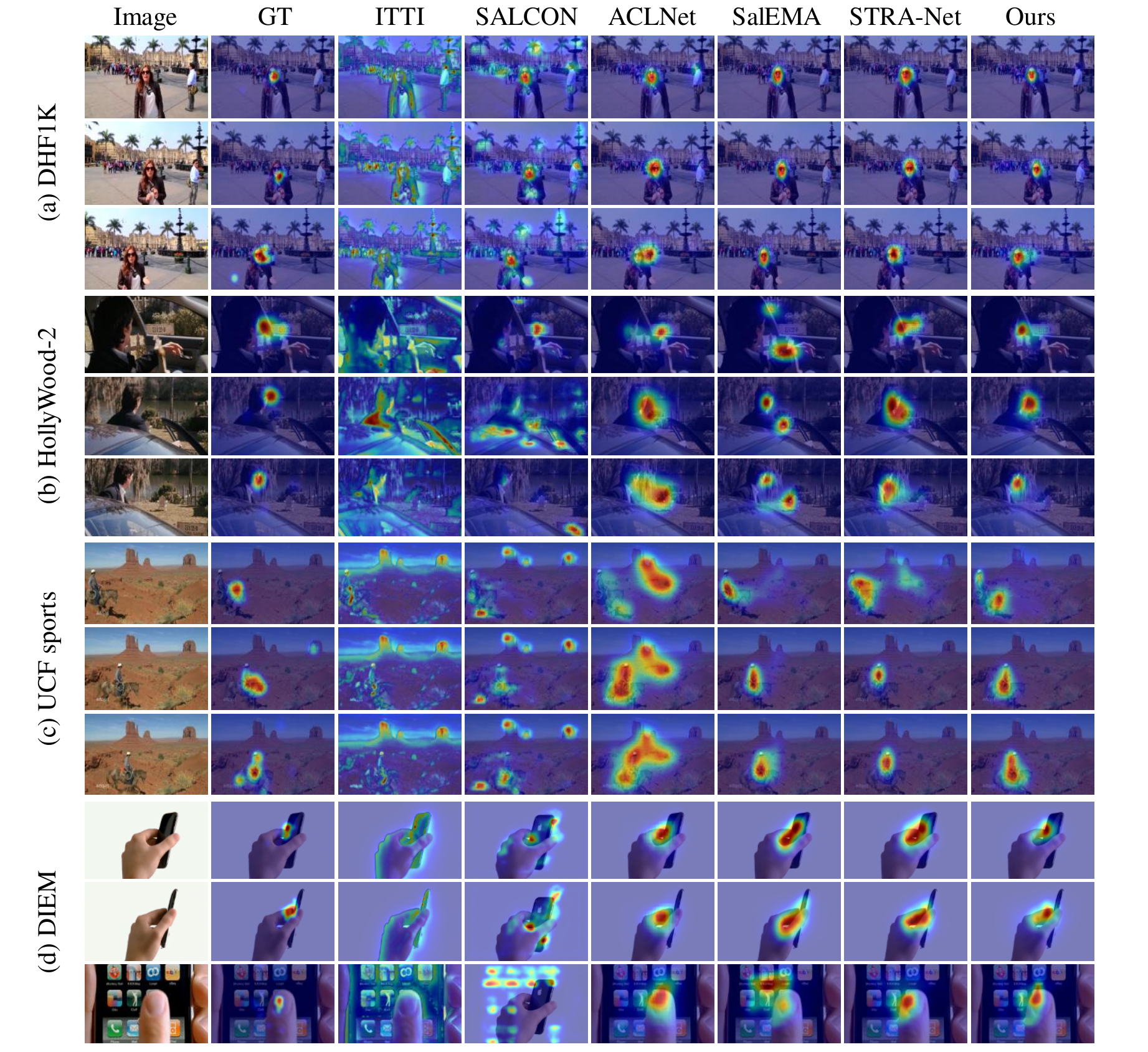}
\caption{Qualitative comparisons with other methods including ITTI~\cite{itti1998model}, SALICON~\cite{huang2015salicon}, ACLNet~\cite{wang2018revisiting}, SalEMA~\cite{linardos2019simple}, STRA-Net~\cite{lai2019video} on (a) DHF1K, (b) HollyWood-2, (c) UCF sports, (d) DIEM. We show one example video with three frames for demonstration purpose on each dataset.}
\label{results}
\end{figure*}

\section{Experiments}
\subsection{Implementation Details}
We adopt the frequently-used VGG16~\cite{simonyan2014very} pre-trained on ImageNet~\cite{krizhevsky2012imagenet} as the backbone network to extract three feature maps with different resolutions.
All the other parameters are trained from scratch except for the backbone network.
%
%Since most training video sequences are densely sampled in real-time, the difference between two adjacent frames is unobtrusive.
The neighbor size of the reference frame is set to $2T+1=5$.
%
%
%For this reason, we fusion the 5 aligned features by concatenation operation and 1 $\times$ 1 convolution for enhancing the feature representation capability of the center frame.
%
%After the Bi-ConvLSTM unit, the final features are sent to decoder, which consists of 1 $\times$ 1 convolution operator and Bilinear interpolation.
%
Each video training batch contains $15$ consecutive frames from the same video with batch size $4$.
We randomly select the video and its starting frame for each training sample.
All the training frames are scaled to $256\times320$ pixels, and the surrounding pixels of the image are padded with $0$ if its size does not match to $256\times320$ pixels.
The ground-truth binary fixation mask and its continuous attention map are scaled to $128\times160$ pixels.
We use the Adam optimizer~\cite{kingma2014adam} to learn the whole network parameters $\bm{\theta}_{\mathrm{CNN}}$ in an end-to-end manner without any post processing.
The learning rate is set to $1e-7$ and the model converges at about $20,000$ steps.
The proposed model is implemented in PyTorch and one Nvidia RTX 2080Ti GPU is used for acceleration.
The whole training process takes about $15$ hours.

\renewcommand\arraystretch{1}
\begin{table*}[t]
\caption{{Quantitative results on DHF1K, HollyWood-2 and UCF sports} dataset. The best-performing scores are marked in \textbf{bold}, and the second-best performing scores are underlined. ${}^*$ represents the DL-based model.}\smallskip
\centering
\resizebox{1\columnwidth}{!}{
\smallskip\begin{tabular}{c|r|c|c|c|c|c||c|c|c|c|c||c|c|c|c|c}
\toprule[1pt]
\multirow{2}{*}{} &\multirow{2}{*}{\diagbox{Methods}{Dataset}} &\multicolumn{5}{c||}{\textbf{DHF1K}} &\multicolumn{5}{c||}{\textbf{HollyWood-2}} &\multicolumn{5}{c}{\textbf{UCF sports}} \\
\cline{3-17}
& &AUC-J$\uparrow$ &SIM$\uparrow$ &s-AUC$\uparrow$ &CC$\uparrow$ &NSS$\uparrow$ &AUC-J$\uparrow$ &SIM$\uparrow$ &s-AUC$\uparrow$ &CC$\uparrow$ &NSS$\uparrow$ &AUC-J$\uparrow$ &SIM$\uparrow$ &s-AUC$\uparrow$ &CC$\uparrow$ &NSS$\uparrow$ \\
\hline
\hline
\multirow{6}{*}{\rotatebox{90}{Static Models}}
&ITTI~\cite{itti1998model}                &0.774 &0.162 &0.553 &0.233 &1.207 &0.788 &0.221 &0.607 &0.257 &1.076 &0.847 &0.251 &0.725 &0.356 &1.640 \\
&GBVS~\cite{harel2007graph}               &0.828 &0.186 &0.554 &0.283 &1.474 &0.837 &0.257 &0.633 &0.308 &1.336 &0.859 &0.274 &0.697 &0.396 &1.818 \\
&${}^*$SALICON~\cite{huang2015salicon}    &0.857 &0.232 &0.590 &0.327 &1.901 &0.856 &0.321 &0.711 &0.425 &2.013 &0.848 &0.304 &0.738 &0.375 &1.838 \\
&${}^*$Shallow-Net~\cite{pan2016shallow}  &0.833 &0.182 &0.529 &0.295 &1.509 &0.851 &0.276 &0.694 &0.423 &1.680 &0.846 &0.276 &0.691 &0.382 &1.789 \\
&${}^*$Deep-Net~\cite{pan2016shallow}     &0.855 &0.201 &0.592 &0.331 &1.775 &0.884 &0.300 &0.736 &0.451 &2.066 &0.861 &0.282 &0.719 &0.414 &1.903 \\
&${}^*$DVA~\cite{wang2017deep}            &0.860 &0.262 &0.595 &0.358 &2.013 &0.886 &0.372 &0.727 &0.482 &2.459 &0.872 &0.339 &0.725 &0.439 &2.311 \\
\hline
\hline
\multirow{13}{*}{\rotatebox{90}{Dynamic Models}}
&PQFT~\cite{guo2009novel}                       &0.699 &0.139 &0.562 &0.137 &0.749 &0.723 &0.201 &0.621 &0.153 &0.755 &0.825 &0.250 &0.722 &0.338 &1.780 \\
&Seo \textit{et al.}~\cite{seo2009static}       &0.635 &0.142 &0.499 &0.070 &0.334 &0.652 &0.155 &0.530 &0.076 &0.346 &0.831 &0.308 &0.666 &0.336 &1.690 \\
&Rudoy \textit{et al.}~\cite{rudoy2013learning} &0.769 &0.214 &0.501 &0.285 &1.498 &0.783 &0.315 &0.536 &0.302 &1.570 &0.763 &0.271 &0.637 &0.344 &1.619 \\
&Hou \textit{et al.}~\cite{hou2009dynamic}      &0.726 &0.167 &0.545 &0.150 &0.847 &0.731 &0.202 &0.580 &0.146 &0.684 &0.819 &0.276 &0.674 &0.292 &1.399 \\
&Fang \textit{et al.}~\cite{fang2014video}      &0.819 &0.198 &0.537 &0.273 &1.539 &0.859 &0.272 &0.659 &0.358 &1.667 &0.845 &0.307 &0.674 &0.395 &1.787 \\
&OBDL~\cite{hossein2015many}                    &0.638 &0.171 &0.500 &0.117 &0.495 &0.640 &0.170 &0.541 &0.106 &0.462 &0.759 &0.193 &0.634 &0.234 &1.382 \\
&AWS-D~\cite{leboran2016dynamic}                &0.703 &0.157 &0.513 &0.174 &0.940 &0.694 &0.175 &0.637 &0.146 &0.742 &0.823 &0.228 &0.750 &0.306 &1.631 \\
&${}^*$OM-CNN~\cite{jiang2018deepvs}            &0.856 &0.256 &0.583 &0.344 &1.911 &0.887 &0.356 &0.693 &0.446 &2.313 &0.870 &0.321 &0.691 &0.405 &2.089 \\
&${}^*$Two-stream~\cite{bak2017spatio}          &0.834 &0.197 &0.581 &0.325 &1.632 &0.863 &0.276 &0.710 &0.382 &1.748 &0.832 &0.264 &0.685 &0.343 &1.753 \\
&${}^*$ACLNet~\cite{wang2018revisiting}         &0.890 &0.315 &0.601 &0.434 &2.354 &0.913 &\textbf{0.542} &0.757 &0.623 &3.086 &0.905 &0.496 &0.767 &0.603 &3.200 \\
&${}^*$SalEMA~\cite{linardos2019simple}         &0.890 &\textbf{0.465} &0.667 &0.449 &2.573 &0.919 &0.487 &0.708 &0.613 &3.186 &0.906 &0.431 &0.740 &0.544 &2.638 \\
&${}^*$TASED-Net~\cite{min2019tased}            &0.895 &0.361 &\textbf{0.712} &0.470 &2.667 &0.918 &0.507 &0.768 &0.646 &3.302 &0.899 &0.469 &0.752 &0.582 &2.920 \\
&${}^*$STRA-Net~\cite{lai2019video}             &0.895 &0.355 &0.663 &0.458 &2.558 &0.923 &0.536 &0.774 &\underline{0.662} &3.478 &0.914 &\textbf{0.535} &\underline{0.790} &\textbf{0.645} &\underline{3.472} \\
\hline
\hline
\multirow{4}{*}{\rotatebox{90}{Ours}}
&Training setting (i)   &\underline{0.896} &\underline{0.390} &0.679 &\textbf{0.479} &\textbf{2.758} &0.915 &0.504 &0.786 &0.613 &3.461 &0.896 &0.455 &0.760 &0.558 &2.985 \\
&Training setting (ii)  &0.888 &0.309 &0.670 &0.438 &2.479 &\textbf{0.934} &0.529 &\textbf{0.806} &\textbf{0.672} &\textbf{3.936} &0.913 &0.418 &0.753 &0.566 &3.039 \\
&Training setting (iii) &0.851 &0.260 &0.664 &0.327 &1.876 &0.893 &0.402 &0.752 &0.481 &2.627 &\textbf{0.921} &\underline{0.497} &\textbf{0.799} &\underline{0.612} &\textbf{3.676} \\
&Training setting (iv)  &\textbf{0.900} &0.353 &\underline{0.680} &\underline{0.476} &\underline{2.685} &\underline{0.928} &\underline{0.537} &\underline{0.800} &0.661 &\underline{3.804} &\underline{0.917} &0.494 &0.785 &0.599 &3.406 \\
\hline
\hline

\end{tabular}
}
\label{leaderboard}
\end{table*}

\subsection{Evaluation Datasets}
We conduct extensive evaluations on four widely-used eye-tracking benchmark datasets.

\textbf{DHF1K}~\cite{wang2019revisiting}: It consists of $1,000$ high-quality elaborately-selected video sequences that are annotated by $17$ observers using an eye tracker device.
The videos therein have diverse contents, varied motion patterns, various objects, large scale and high quality.
The videos are divided into a training set of $600$ videos and a validation set of $100$ videos that are publicly available, but the fixation labels of the remaining $300$ videos are not released, which are used to validate the generalization capability of the model fairly.

\textbf{HollyWood-2}~\cite{mathe2014actions}: It contains $1,707$ videos selected from the HollyWood-2 action recognition dataset~\cite{marszalek2009actions} that is collected from a set of $69$ HollyWood movies.
$12$ action classes are included such as \textit{hugging}, \textit{kissing}, \textit{running}, \textit{etc}.
The whole data consists of a training set of $823$ sequences and a test set of $884$ sequences.
It is one of the largest and most challenging available dataset for VSP.

\textbf{UCF Sports}~\cite{mathe2014actions}: It includes $150$ videos from the UCF sports action datasets~\cite{rodriguez2008action}, which covers $9$ sports action classes such as \textit{swinging}, \textit{lifting}, \textit{skateboarding}, \textit{etc}.
The dataset consists of $103$ videos for training and $47$ videos for testing.

\textbf{DIEM}~\cite{mital2011clustering}: It has $84$ videos that are collected from $50$ participants which are widely used for studying human-eye fixation attention.
Following~\cite{hossein2015many,lai2019video}, we select the same $20$ videos as the testing set.

For fair comparison, we leverage the standard training strategy in~\cite{wang2018revisiting}, which consists of $4$ training settings with the training sets of (i) DHF1K, (ii) HollyWood-2, (iii) UCF sports, (iv) DHF1K + HollyWood-2 + UCF sports.
Meanwhile, we use the testing sets of DHF1K, HollyWood-2 and UCF sports to evaluate the performance.
Furthermore, to further assess the generalization capability of our model, we evaluate it on the testing set of DIEM which has no training set available.

\renewcommand\arraystretch{0.7}
\begin{table*}[t]
\caption{{Quantitative results on DIEM} dataset ({without training}). The best-performing scores are marked in \textbf{bold}. The second best-performing scores are underlined. ${}^*$ represents the DL-based model.}\smallskip
\centering
\resizebox{1\columnwidth}{!}{
\smallskip\begin{tabular}{c|r|c|c|c|c|c}
\toprule[1pt]
&Methods &AUC-J$\uparrow$ &SIM$\uparrow$ &s-AUC$\uparrow$ &CC$\uparrow$ &NSS$\uparrow$ \\
\hline

\multirow{6}{*}{\rotatebox{90}{Static Models}}
&ITTI~\cite{itti1998model}                &0.791 &0.132 &0.653 &0.196 &1.103 \\
&GBVS~\cite{harel2007graph}               &0.813 &0.156 &0.633 &0.214 &1.198 \\
&${}^*$SALICON~\cite{huang2015salicon}    &0.793 &0.171 &0.674 &0.270 &1.650 \\
&${}^*$Shallow-Net~\cite{pan2016shallow}  &0.838 &0.188 &0.620 &0.297 &1.646 \\
&${}^*$Deep-Net~\cite{pan2016shallow}     &0.849 &0.164 &0.697 &0.291 &1.650 \\
&${}^*$DVA~\cite{wang2017deep}            &0.868 &0.237 &\textbf{0.721} &0.386 &2.347 \\
\hline
\hline

\multirow{11}{*}{\rotatebox{90}{Dynamic Models}}
&PQFT~\cite{guo2009novel}                             &0.724 &0.126 &0.649 &0.144 &0.856 \\
&Seo \textit{et al.}~\cite{seo2009static}             &0.723 &0.130 &0.568 &0.116 &0.665 \\
&Rudoy \textit{et al.}~\cite{rudoy2013learning}       &0.775 &0.150 &0.618 &0.260 &1.390 \\
&Hou \textit{et al.}~\cite{hou2009dynamic}            &0.735 &0.142 &0.589 &0.128 &0.735 \\
&Fang \textit{et al.}~\cite{fang2014video}            &0.823 &0.167 &0.636 &0.251 &1.423 \\
&OBDL~\cite{hossein2015many}                          &0.762 &0.165 &0.694 &0.221 &1.289 \\
&AWS-D~\cite{leboran2016dynamic}                      &0.774 &0.150 &0.695 &0.216 &1.252 \\
&${}^*$OM-CNN~\cite{jiang2018deepvs}                  &0.857 &0.238 &0.693 &0.371 &2.235 \\
&${}^*$Two-stream~\cite{bak2017spatio}                &0.859 &0.256 &0.682 &0.366 &2.171 \\
&${}^*$ACLNet~\cite{wang2018revisiting}               &0.881 &0.277 &0.693 &0.396 &2.368 \\
&${}^*$STRA-Net~\cite{lai2019video}                   &0.870 &0.306 &0.678 &0.408 &\textbf{2.452} \\

\hline
\hline

\multirow{4}{*}{\rotatebox{90}{Ours}}
&Training setting (i)   &\underline{0.880} &\underline{0.391} &\underline{0.718} &\underline{0.472} &\underline{2.346} \\
&Training setting (ii)  &0.870 &0.355 &0.698 &0.468 &2.359 \\
&Training setting (iii) &0.831 &0.296 &0.676 &0.351 &1.828 \\
&Training setting (iv)  &\textbf{0.889} &\textbf{0.396} &0.711 &\textbf{0.490} &\underline{2.346} \\
\hline
\hline

\end{tabular}
}
\label{DIEM}
\end{table*}

\subsection{Comparison Results}
As~\cite{wang2018revisiting,linardos2019simple,min2019tased,lai2019video}, we use the widely-used evaluation metrics~\cite{bylinskii2018different} to evaluate the comparative methods, including Normalized Scanpath Saliency (NSS), Similarity (SIM), shuffled AUC (s-AUC), linear Correlation Coefficient (CC) and AUC-J.
%
%More detailed introduction of the four metrics can be found in.

%
We compare the proposed approach with $19$ saliency models, including $6$ static models (ITTI~\cite{itti1998model}, GBVS~\cite{harel2007graph}, SALICON~\cite{huang2015salicon}, Shallow-Net~\cite{pan2016shallow}, Deep-Net~\cite{pan2016shallow}, DVA~\cite{wang2017deep}) and $13$ dynamic models (PQFT~\cite{guo2009novel}, Seo~\textit{et al.}~\cite{seo2009static}, Rudoy~\textit{et al.}~\cite{rudoy2013learning}, Hou~\textit{et al.}~\cite{hou2009dynamic}, Fang~\textit{et al.}~\cite{fang2014video}, OBDL~\cite{hossein2015many}, AWS-D~\cite{leboran2016dynamic}, OM-CNN~\cite{jiang2018deepvs}, Two-stream~\cite{bak2017spatio}, ACLNet~\cite{wang2018revisiting}, SalEMA~\cite{linardos2019simple}, TASED-Net~\cite{min2019tased}, STRA-Net~\cite{lai2019video}). Among them, SALICON~\cite{huang2015salicon}, Shallow-Net~\cite{pan2016shallow}, Deep-Net~\cite{pan2016shallow}, DVA~\cite{wang2017deep}, OM-CNN~\cite{jiang2018deepvs}, Two-stream~\cite{bak2017spatio}, ACLNet~\cite{wang2018revisiting} and STRA-Net~\cite{lai2019video} are the DL-based models while the others are the traditional models.
\paragraph{Results on DHF1K} We test our model on the testing set of DHF1K, which contains $300$ videos without publicly released ground-truth annotations of human eye-tracking maps available. A public server is used to report the results on the test set.
It is a fair and large test set for verifying the generalization capability of our model.
Table~\ref{leaderboard} lists the results of our model in terms of AUC-J, SIM, s-AUC, CC and NSS. Among them, our model achieves the best CC and NSS scores and the second-best SIM and AUC-J scores with training setting (i). Meanwhile, for training setting (iv), our model achieves the best AUC-J score and the second-best s-AUC, CC and NSS scores.
Besides, the results of our model with all the settings are much better than the statistic models.
For the settings (ii) and (iii), our model achieves competing performance among the dynamic models, following the top-performing methods including ACLNet, SalEMA, TASED-Net and STRA-Net.
%
%Some of the methods including STRA-Net~\cite{lai2019video} and TASED-Net~\cite{min2019tased} are the top of the server leaderboard.
%

Figure~\ref{results} (a) shows the qualitative comparison results, where the visual attention maps generated by our model well approach to the ground-truth, which more focus on the saliency targets and are not disturbed by the transitions of the backgrounds.

\paragraph{Results on HollyWood-2} As listed by Table~\ref{leaderboard}, the performance of our approach is superior to the other methods,  especially in training setting (ii), our model achieves the best scores in terms of s-AUC, CC and NSS.
The dataset mainly focuses on task-driven viewing mode whose contents are limited to human actions and movie scenes.
Therefore, training with the same distribution of the training setting yields much better results.
Figure~\ref{results} (b) shows a man opening the door from the car and going out, where our predicted saliency maps can more accurately track the salient objects than the saliency maps generated by the other methods.

\paragraph{Results on UCF sports} Compared to all the other models, the proposed method achieves the best or second-best performance in terms of all metrics under training setting (iii). For the other training settings, our method achieves comparative results compared to the top-performing methods such as ACLNet, SalEMA, TASED-Net and STRA-Net.
Figure~\ref{results} (c) shows the scene where a man rides a horse in the desert.
The salient targets we focus on at different times will change, and our predicted visual maps enable to precisely move the attention shift from horse to man compared to the other methods.

\paragraph{Results on DIEM} For evaluating the generalization capability of our model, we do not use any data of DIEM to train our model.
Following~\cite{hossein2015many, lai2019video}, we evaluate our model on the testing set of DIEM, containing the first $300$ frames of each video.
Table~\ref{leaderboard} lists the quantitative results, which achieves competitive results compared to the other methods, especially under training settings (i) and (iv) where the proposed method achieves the best or second-best performance in terms of almost all metrics.
\renewcommand\arraystretch{0.7}
\begin{table*}[t]
\caption{Results of ablative experiments of the proposed model on the UCF sports testing set. The best scores are marked in \textbf{bold}. The symbol '\textit{w/o.}' denotes removing a specific module.}\smallskip
\centering
\resizebox{1\columnwidth}{!}{
\smallskip\begin{tabular}{l|c|c|c|c|c}
\toprule[1pt]
UCF sports Dataset &AUC-J$\uparrow$ &SIM$\uparrow$ &s-AUC$\uparrow$ &CC$\uparrow$ &NSS$\uparrow$ \\
\hline

Ours                              &\textbf{0.921} &\textbf{0.497} &\textbf{0.799} &\textbf{0.612} &\textbf{3.676} \\
\hline

\textit{w/o.} Bi-ConvLSTM                          &0.903 &0.470 &0.788 &0.571 &3.467 \\
\textit{w/o.} forward ConvLSTM                     &0.907 &0.467 &0.778 &0.590 &3.293 \\
\textit{w/o.} backward ConvLSTM                    &0.894 &0.457 &0.789 &0.536 &3.393 \\
\hline

replace DConv to Conv in MDAN                      &0.905 &0.452 &0.775 &0.549 &3.265 \\
\textit{w/o.} $p5$ in MDAN                           &0.894 &0.442 &0.781 &0.565 &3.483 \\
\textit{w/o.} $p4$ and $p5$ in MDAN                       &0.895 &0.425 &0.776 &0.548 &3.168 \\
\textit{w/o.} $p3$ in MDAN                           &0.900 &0.451 &0.783 &0.543 &3.236 \\
\textit{w/o.} $p3$ and $p4$ in MDAN                       &0.874 &0.423 &0.780 &0.516 &3.357 \\
\hline

\end{tabular}
}
\label{ablation}
\end{table*}

\subsection{Ablative Study}
To further show our main contributions, we compare different variants of our model including those without Bi-ConvLSTM and replacing all Dconv in MDAN to regular convolution.
Table~\ref{ablation} lists the results of ablative experiments on the UCF sports testing set under training setting (iii).
We can observe that without Bi-ConvLSTM, the scores of all the metrics decline to some degree such as the AUC-J score drops from $0.921$ to $0.9034$ and the CC score reduces by $0.04$ from $0.612$ to $0.571$, verifying the effectiveness of Bi-ConvLSTM.
Moreover, replacing Dconv to regular convolution in MDAN causes the performance degradation.
For example, the SIM score drops from $0.497$ to $0.452$ and the NSS score decreases from $3.676$ to $3.265$, demonstrating the effectiveness of using DConv for feature alignment across frames.
Finally, to make further efforts to verify the effectiveness of other modules, we conduct some ablative experiments including: removing forward ConvLSTM, backward ConvLSTM, $p3$, $p4$, $p5$ in MDAN, respectively.
The extra results are also listed in the bottom row of Table~\ref{ablation}, which confidently validate that each component in Bi-ConvLSTM and MDAN has a positive effect to boost the performance of the proposed approach.

\section{Conclusions}
In this paper, we have presented an effective enhanced spatiotemporal alignment network for VSP, mainly including two novel module designs: the MDAN and the Bi-ConvLSTM.
The MDAN makes multi-resolution feature alignment between the reference and its neighboring frames in a coarse-to-fine manner, which is good at handling various motions.
Afterwards, the output features of MDAN are further enhanced by the Bi-ConvLSTM that fully captures the long-time temporal information in both forward and backward timing directions.
Extensive experiments on four VSP benchmark datasets have clearly demonstrated superiority of the proposed method to state-of-the-art methods in terms of five metrics.

\bibliography{pr-cj}

\end{document}